\documentclass[letterpaper, 10pt, journal]{IEEEtran}
\usepackage[centertags]{amsmath}
\usepackage{amssymb,amsthm}
\usepackage{enumitem}
\usepackage{graphicx,subfigure}
\usepackage{siunitx}
\usepackage{algorithm}
\usepackage{algorithmic}
\usepackage[hidelinks]{hyperref}
\usepackage{cleveref}
\usepackage{booktabs,multirow,xspace}
\allowdisplaybreaks
\graphicspath{{./Figs/}}

\title{\LARGE \bfseries
    Constrained Imitation Learning for\\ a Flapping Wing Unmanned Aerial Vehicle
}

\author{Tejaswi K. C. and Taeyoung Lee%
	\vspace{-0.1in}
    \thanks{Tejaswi K. C. and Taeyoung Lee, Mechanical and Aerospace Engineering, The George Washington University, Washington DC 20052 {\tt kctejaswi999@gmail.com,tylee@gwu.edu}}%
    \thanks{\textsuperscript{\footnotesize\ensuremath{*}}This research has been supported in part by NSF under the grants NSF CMMI-1761618 and CMMI-1760928.}
}

\DeclareMathOperator*{\argmin}{arg\,min}

\newcommand{\scalefont}[1]{\scalebox{1.5}{#1}}

\newcommand{\norm}[1]{\ensuremath{\left\| #1 \right\|}}
\newcommand{\bracket}[1]{\ensuremath{\left[ #1 \right]}}
\newcommand{\braces}[1]{\ensuremath{\left\{ #1 \right\}}}
\newcommand{\parenth}[1]{\ensuremath{\left( #1 \right)}}

\newcommand{\trs}[1]{\mathrm{tr}\ensuremath{[#1]}}

\newcommand{\SO}[1]{\ensuremath{\mathsf{SO(#1)}}}
\newcommand{\so}[1]{\ensuremath{\mathfrak{so}(#1)}}

\renewcommand{\Re}{\ensuremath{\mathbb{R}}}

\newcommand{\ad}{\ensuremath{\mathrm{ad}}}

\newcommand{\G}{\ensuremath{\mathsf{G}}}
\newcommand{\g}{\ensuremath{\mathfrak{g}}}

\newcommand{\MATLAB}{\textsc{Matlab}\xspace}

\begin{document}

\maketitle

\begin{abstract}
This paper presents a data-driven optimal control policy for a micro flapping wing unmanned aerial vehicle. 
First, a set of optimal trajectories are computed off-line based on a geometric formulation of dynamics that captures the nonlinear coupling between the large angle flapping motion and the quasi-steady aerodynamics. 
Then, it is transformed into a feedback control system according to the framework of imitation learning. 
In particular, an additional constraint is incorporated through the learning process to enhance the stability properties of the resulting controlled dynamics. 
Compared with conventional methods, the proposed constrained imitation learning eliminates the need to generate additional optimal trajectories on-line, without sacrificing stability. 
As such, the computational efficiency is substantially improved. 
Furthermore, this establishes the first nonlinear control system that stabilizes the coupled longitudinal and lateral dynamics of flapping wing aerial vehicle without relying on averaging or linearization. 
These are illustrated by numerical examples for a simulated model inspired by Monarch butterflies. 
\end{abstract}

\section{Introduction}

Flapping wing aerial vehicles exhibit substantial advantages in energy efficiency and agility, compared against the conventional fixed or rotary wing types whose lift-to-drag ratio deteriorates rapidly as its size is reduced. 
As such, the flapping wing mechanism has been envisaged to be a critical component for micro autonomous drones of the next generation~\cite{floreano2015science}.
However, the corresponding development for control is limited, especially compared with the recent progress in autonomous unmanned aerial vehicles such as quadrotors. 
This is mainly because the flapping wing aerial vehicles correspond to infinite-dimensional nonlinear time-varying systems, where unsteady aerodynamics is coupled with structural deformation of wings and body dynamics in a sophisticated manner.

Most of the existing control systems for flapping wing aerial vehicles bypass these challenges by averaging the linearized dynamics over a flapping period~\cite{Orlowski2012,sun2014insect,shyy2016aerodynamics,Deng2006}.
These works exploit the large disparity in time scales of wingbeat frequency and flight dynamics by utilizing high frequency oscillations of small wings.
Consequently, they are not suitable for flapping wing aerial vehicles or insect flight with a relatively large wing flapping at a low frequency, such as in Monarch butterflies~\cite{Kang2018Experimental,Sridhar_2021},
which exhibit remarkable flight characteristics migrating over three thousands miles disproportionate to their size.

Recently, the authors have proposed a geometric model for flapping wing aerial vehicles inspired by Monarch, where rigid bodies corresponding to thorax, abdomen, and two wings are interconnected via spherical joints~\cite{csribb21}.
This is formulated in an intrinsic fashion as a Lagrangian/Hamiltonian system evolving on the configuration space of a Lie group, such that any arbitrary maneuver involving large-angle flapping is described globally without singularities caused by local parameterization. 
Based on this, a feedback controller is developed for the longitudinal dynamics, where controller parameters are optimized to guarantee stability according to the Lyapunov-Floquet theory~\cite{CKangPACC21}.
This work is restricted to the pitching dynamics and the translations confined in a vertical plane. 
Next, optimal control trajectories are constructed via direct optimization for the coupled dynamics of the longitudinal mode and the lateral mode~\cite{CLeeIFAC21}.
While this approach deals with arbitrary maneuvers in the three-dimensional space, it is computationally expensive due to the extensive number of iterations involved, and therefore, it can not be implemented in real time.
In short, due to the complexities of the dynamics, it is challenging to design a control system from Lyapunov stability analysis or model-predictive online optimization.

To overcome these challenges in aerial robotic applications, MPC implementation has been combined with Guided Policy Search to train neural network policies which are a lot faster~\cite{zhang2016learning}.
Further related work include an adaptive MPC teacher which produces actions to optimize the cost at hand as well as the learned network policy~\cite{kahn2017plato}.
There has been progress even in real world implementations of such sensorimotor neural networks trained from available data.
Directly mapping states to control values, \cite{li2020aggressive} propose offline training using trajectories generated from optimal control.
Extreme acrobatic maneuvers have been performed using an end-to-end framework on a real quadrotor with limited sensor measurements and computation~\cite{kaufmann2020deep}.
More recently, similar policies trained in simulation have been utilized for controlling flight in challenging environments which could even include obstacles~\cite{loquercio2021learning}.

A common theme here is to construct a control policy by emulating the behavior of an expert, which could be the optical controller.
This idea is based upon imitation learning, where a set of ideal control actions demonstrated by the expert is reproduced via supervised learning with the goal of mimicking the expert control policy~\cite{argall2009survey}.
A naive implementation of imitation learning, where a neural network is trained to model expert demonstrations, is referred to as behavior cloning~\cite{Batavia-1996-14235}.
However, the performance of behavior cloning is not satisfactory in practice, as the mismatch of distributions between offline training data and online trajectories accumulates over time.
To address this, it has been proposed to repeatedly augment the training data with the actual state encountered during online implementation and the corresponding optimal control input of the expert, which is referred to as the DAgger algorithm~\cite{ross2011reduction}.
But it has been shown that this can lead to visiting potentially unsafe states which can be avoided by seeking supervisor help~\cite{zhang2016query}.
More recent approaches include DART, which injects noise into the expert itself so that the trained model is robust~\cite{laskey2017dart}.
However, these require that the expert is continuously queried, and also, the neural network should be retrained for training data that are constantly renewed and enlarged. 

In this paper, to address these challenges of imitation learning for flapping wing aerial vehicles, we propose a new scheme referred to as constrained imitation learning (COIL). 
The fundamental idea is that we impose an equality constraint on the neural network to guarantee zero control for zero state error, instead of augmenting the training data with additional trajectories. 
This injection of domain knowledge in the form of constraints has been illustrated to improve the performance of learned networks~\cite{borghesi2020improving}.
Various such End-to-end Constrained Optimization Learning methods which deal with integration of optimization into deep learning are surveyed in~\cite{kotary2021end}.
One such approach is the strategic adjustment of targets instead of handling the constraint directly~\cite{detassis2020teaching}, which alleviates computational impediments in constrained supervised learning.
The proposed COIL algorithm obtains inspiration from a modified procedure that has been presented in ~\cite{tejaswi2022iterative}.
However, unlike \cite{tejaswi2022iterative} which deals with output constraints, the constraint that is explicitly enforced here is on the parameters of the neural network.
Hence it is necessary to make additional changes which are discussed in detail in the corresponding \Cref{sec:COIL}. 

The unique advantages of COIL are summarized as follows:
\begin{enumerate}
	\item Imitation learning is tailored for control of dynamic systems to improve stability properties without causing computational burden;
	\item The issue of instability in behavior cloning is addressed without enlarging the training data set;
	\item There is no need to communicate with the expert online during the learning process, which enables imitation learning even when the expert is not available for interaction.
\end{enumerate}

The proposed COIL scheme is applied to the dynamics of flapping wing aerial vehicle, where the optimal control trajectories constructed offline via direct optimization are considered as expert demonstration. 
In other words, the goal is to construct a data-driven feedback control system mimicking optimal behaviors.
The above third property of COIL is particularly useful as there is no need to solve time-consuming, offline trajectory optimization repeatedly. 
Further, this overcomes the aforementioned challenges in designing control systems for flapping wing aerial vehicles, as it avoids on-policy optimization required in model predictive control.
Nor does it require Lyapunov stability analysis for sophisticated equations of motion involving aerodynamic forces and moments. 

This paper is organized as follows. 
The dynamics of a flapping wing aerial vehicle and the offline optimal control framework are presented in \Cref{sec:dynamics}.
The COIL scheme is proposed in \Cref{sec:IL} and it is verified numerically in \Cref{sec:results}.
An open-source \MATLAB implementation of the dynamical model, optimal controller and learning schemes is available in~\cite{cleegithub21}.

\section{Preliminaries}\label{sec:dynamics}

In this section, we summarize the dynamic model of a flapping wing unmanned aerial vehicle (FWUAV) inspired by Monarch butterfly~\cite{sridhar2020geometric,csribb21}.
The equations of motion and wing kinematics are formulated, and a geometric numerical integration scheme is presented.

Throughout this paper, the three-dimensional special orthogonal group is denoted by $ \SO3 = \lbrace R \in \mathbb{R}^{3\times3} \mid R^T R = I, \det(R) = 1 \rbrace $, 
and its Lie algebra is $ \so3 = \lbrace A \in \mathbb{R}^{3\times3} \mid A = -A^T \rbrace $.
The \textit{hat} map $ \wedge : \mathbb{R}^3 \to \so3 $ is defined such that $\hat x y = x\times y$ for any $x,y\in\Re^3$,
and its inverse is the \textit{vee} map, $ \vee:\so3\rightarrow\Re^3 $.
The matrix $e_i\in\Re^n$ denotes the $i$-th standard basis of $\Re^n$ for an appropriate dimension $n$, e.g., $e_1=(1,0,\ldots, 0)\in\Re^n$.
For all numerical examples, the units are in $\si{kg}$, $\si{m}$, $\si{s}$, and $\si{rad}$, unless specified otherwise.

\subsection{Multiple Rigid Body Formulation}
Let the inertial frame be $\mathcal{F}_I=\{\mathbf{i}_x,\mathbf{i}_y,\mathbf{i}_z\}$.
We model FWUAV as a composition of three rigid bodies, namely thorax and two wings, interconnected by spherical joints.
The motion of each component is described as follows.

\setlength{\unitlength}{0.1\textwidth}
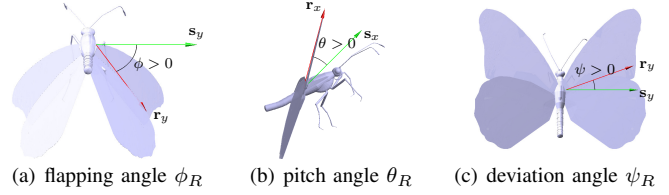
\begin{figure}
        \footnotesize
        \subfigure[flapping angle $\phi_R$]{
            \scalebox{0.475}{
                \begin{picture}(3,2.6)(0,0)
                    \put(0,0){\includegraphics[trim={4cm 3cm 4cm 1.5cm},clip,width=0.3\textwidth]{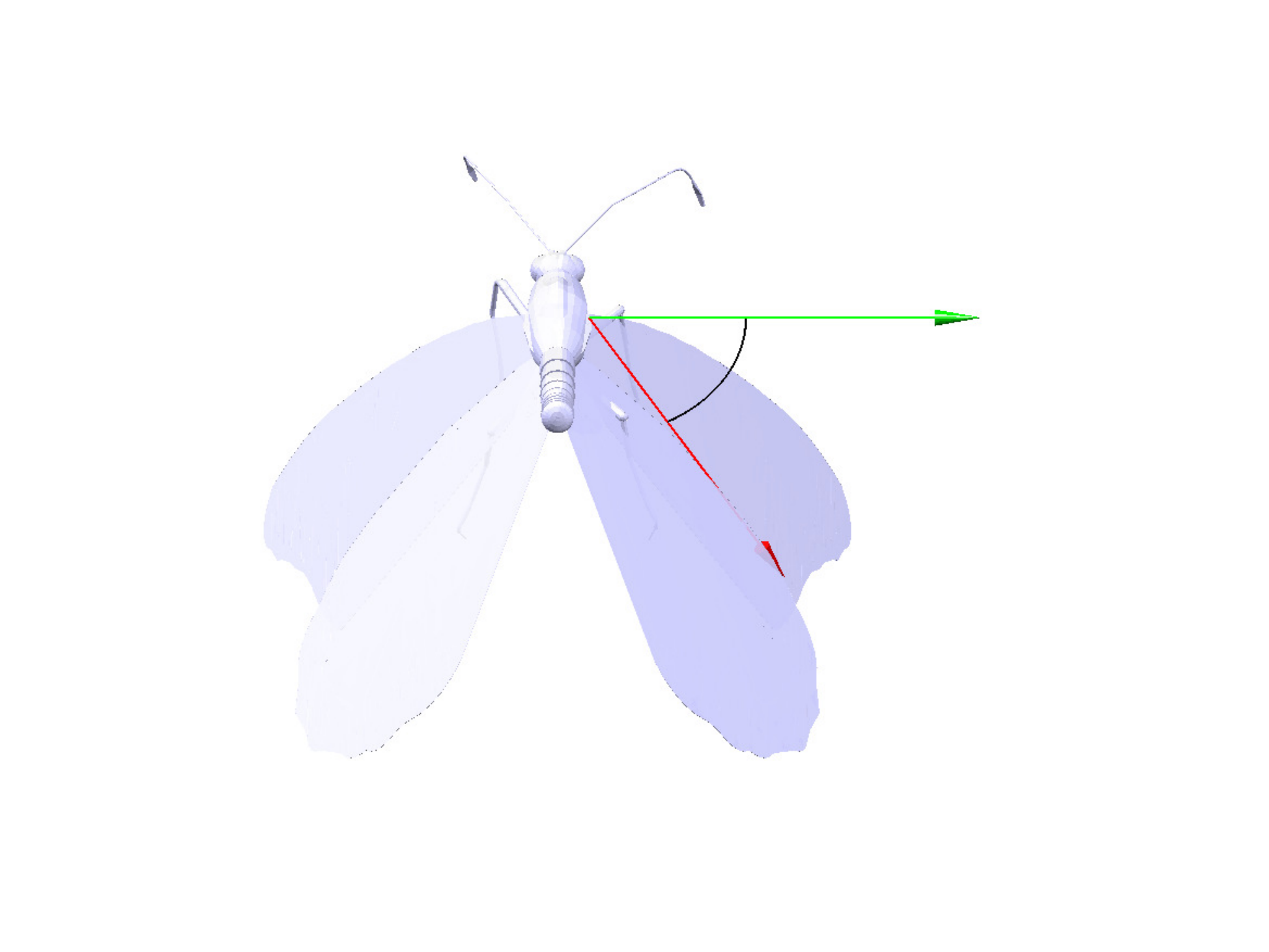}}
                    \put(1.9,1.4){\scalefont{$\phi>0$}}
                    \put(2.75,1.9){\scalefont{$\mathbf{s}_y$}}
                    \put(2.2,0.6){\scalefont{$\mathbf{r}_y$}}
                \end{picture}
            }
        }
        \subfigure[pitch angle $\theta_R$]{
            \scalebox{0.475}{
                \begin{picture}(3,2.6)(0,0)
                    \put(0,0){\includegraphics[trim={4cm 3cm 2.5cm 1.5cm},clip,width=0.3\textwidth]{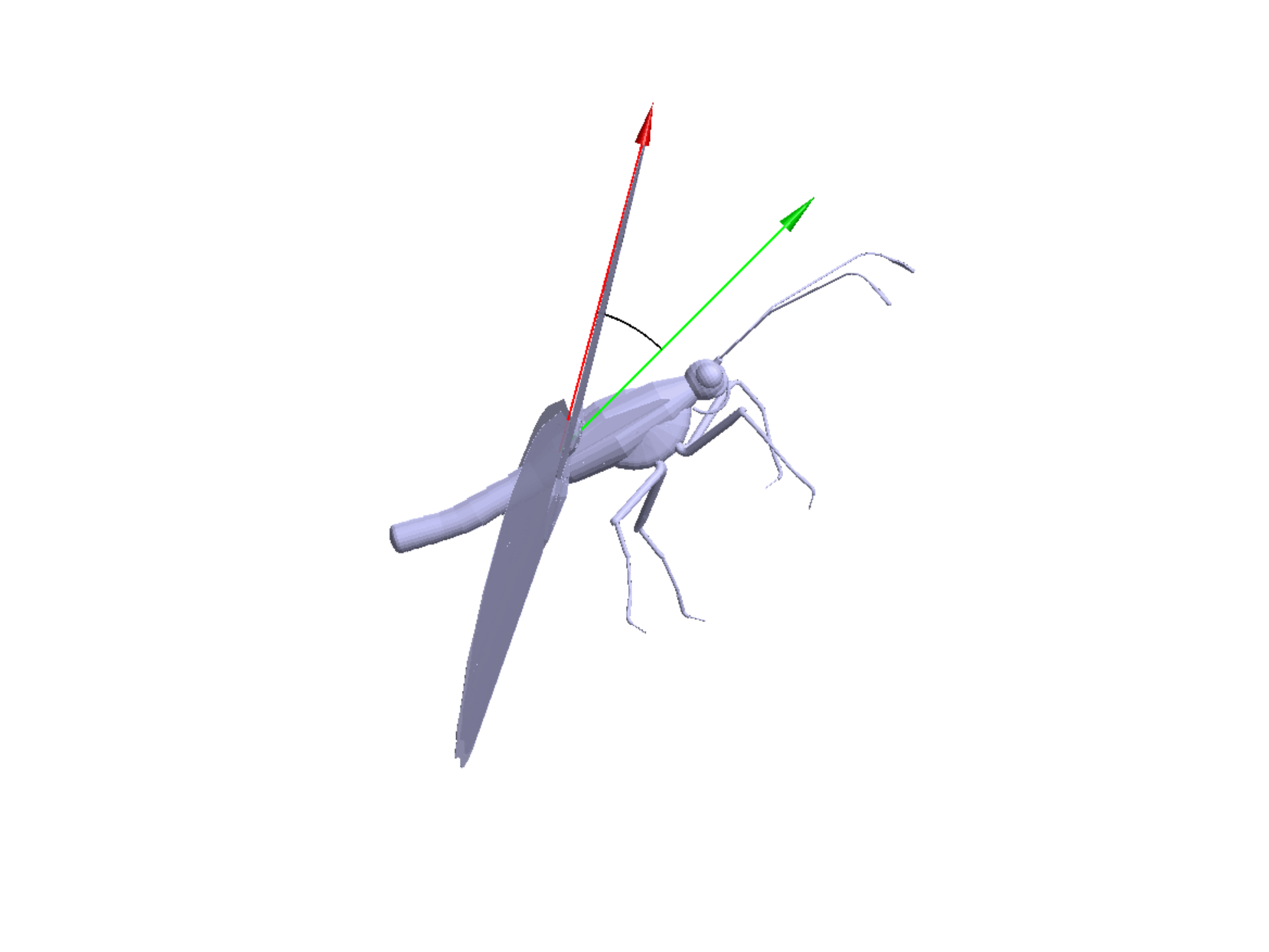}}
                    \put(1.3,1.7){\scalefont{$\theta>0$}}
                    \put(1.1,2.3){\scalefont{$\mathbf{r}_x$}}
                    \put(2.0,1.85){\scalefont{$\mathbf{s}_x$}}
                \end{picture}
            }
        }
        \subfigure[deviation angle $\psi_R$]{
            \scalebox{0.475}{
                \begin{picture}(3,2.6)(0,0)
                    \put(0,0){\includegraphics[trim={3cm 2cm 3cm 2cm},clip,width=0.3\textwidth]{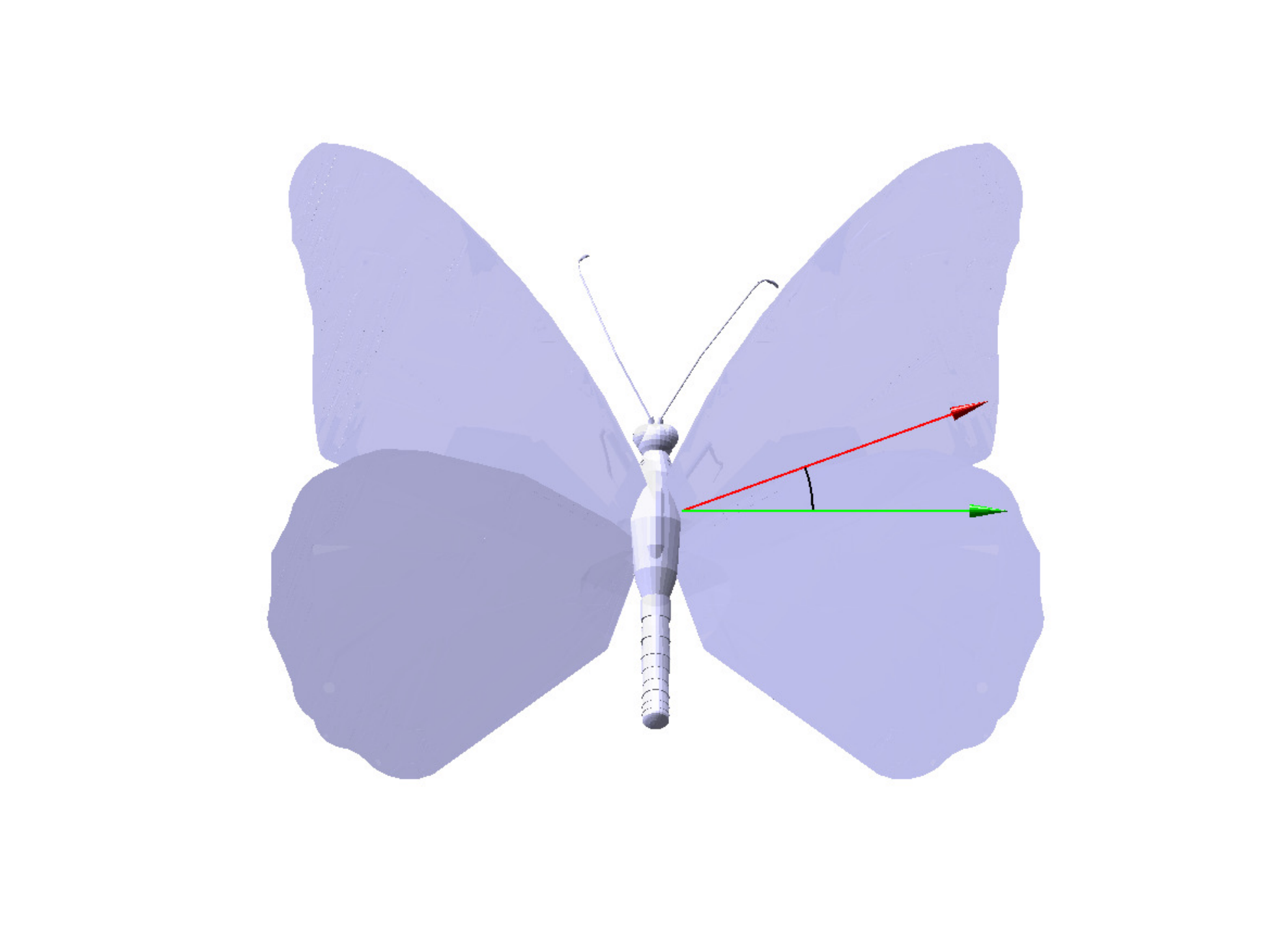}}
                    \put(1.8,1.3){\scalefont{$\psi>0$}}
                    \put(2.8,1.4){\scalefont{$\mathbf{r}_y$}}
                    \put(2.8,0.95){\scalefont{$\mathbf{s}_y$}}
                \end{picture}
            }
        }
    \caption{Euler angles corresponding to right wing attitude~\cite{sridhar2020geometric}
    }\label{fig:wing_Euler}
\end{figure}

\begin{itemize}
	\item \textit{Body:}

	The body-fixed frame $\mathcal{F}_B=\{\mathbf{b}_x,\mathbf{b}_y,\mathbf{b}_z\}$ is fixed at its center of mass whose position is denoted as $ x \in \Re^3 $ in $ \mathcal{F}_I $ and attitude is given by $R\in\SO3$.
	This attitude evolves as $ \dot R = R \hat \Omega, $ where $\Omega\in\Re^3$ is its angular velocity resolved in $\mathcal{F}_B$.

	\item \textit{Right wing:}
	$\mathcal{F}_R=\{\mathbf{r}_x,\mathbf{r}_y,\mathbf{r}_z\}$ is fixed to the right wing at its root and located at a distance $\mu_R\in\Re^3$ in $\mathcal{F}_B$.
	The attitude of $ \mathcal{F}_R $ with respect to an intermediate stroke frame (with the stroke angle, $ \beta $) is characterized by 1--3--2 Euler angles $(\phi_R(t), \psi_R(t), \theta_R(t))$ (see Figure \ref{fig:wing_Euler}).
	Thus its attitude relative to $ \mathcal{F}_B $ is formulated as,
	$
	Q_R= \exp(\beta \hat e_2)\exp(\phi_R \hat e_1) \exp(-\psi_R \hat e_3) \exp(\theta_R\hat e_2),
	$
	with $ \dot Q_R = Q_R \hat \Omega_R $ for $\Omega_R\in\Re^3$. 

	\item \textit{Left Wing:}
	Since $\mathcal{F}_L=\{\mathbf{l}_x, \mathbf{l}_y, \mathbf{l}_z\}$ is defined symmetrically to the right wing,
	$ Q_L = \exp(\beta \hat e_2)\exp(-\phi_L \hat e_1) \exp(\psi_L \hat e_3) \exp(\theta_L \hat e_2) ,$
    and $ \dot Q_L = Q_L \hat \Omega_L $ for $\Omega_L\in\Re^3$.

\end{itemize}

The detailed definition of these frames are presented in~\cite{csribb21}, with an additional component of abdomen.

\subsection{Equations of motion}

The complete configuration of the model can be written as $g=(x,R,Q_R,Q_L)$ that evolves on the Lie group $\G=\Re^3\times \SO3^3$.
Its Lie algebra is $\g = \Re^3 \times \so3^3 \simeq \Re^3 \times (\Re^3)^3$ to which the velocities $\xi = (\dot x, \Omega, \Omega_R, \Omega_L) $ belong.
The corresponding Euler-Lagrange equations have been developed in~\cite{csribb21, tejaswi2021geometric}.

In this paper, we are interested in influencing the motion of FWUAV by controlling the wings, 
assuming that flapping motion of the wing, represented by $Q_R(t), Q_L(t)$, is given as a function of time. 
Thus the whole configuration can be decomposed into a freely varying part and a prespecified part,
\begin{gather}
	g_1 = (x, R), \quad \xi_1 = [\dot x, \Omega], \label{eqn:g1xi1}\\
	g_2 = (Q_R, Q_L), \quad \xi_2 = [\Omega_R, \Omega_L].\label{eqn:g2xi2}
\end{gather}
with  $g=(g_1,g_2)$ and $\xi=(\xi_1,\xi_2)$.
For given $(g_2(t), \xi_2(t))$, the governing equations for $(g_1,\xi_1)$ are given by
\begin{align}
	\dot \xi_1 &= (\mathbf{J}_{11}-C\mathbf{J}_{21})^{-1} \left[ (\ad^*_{\xi_1}\mathbf{J}_{11}-C\ad^*_{\xi_2} +\mathbf{J}_{21} )\xi_1 \right. \nonumber\\
	& \qquad - (\mathbf{L}_{11}-C\mathbf{L}_{21})\xi_1 - (\mathbf{J}_{12}-C\mathbf{J}_{22})\dot \xi_2 \nonumber \\
	& \qquad +(\ad^*_{\xi_1}\mathbf{J}_{12}-C\ad^*_{\xi_2} \mathbf{J}_{22} )\xi_2 - (\mathbf{L}_{12}-C\mathbf{L}_{22})\xi_2 \nonumber \\ 
	&\qquad \left. + \mathbf{f}_{a_1}+\mathbf{f}_{g_1}-C(\mathbf{f}_{a_2}+\mathbf{f}_{g_2}) \right],\label{eqn:EL_xR}
\end{align}
where $\mathbf{J}$ and $\mathbf{L}$ correspond to the inertia matrix and its derivatives, defined by the mass properties of FWUAV and the configuration,
and $\mathbf{f}_a$ and $\mathbf{f}_g$ represent the effects of aerodynamic forces and gravity, respectively. 
Next, $\ad^*$ denotes the coadjoint action of the Lie algebra that can be represented by the matrices
\begin{gather*}
    \ad^*_{\xi_1} = \mathrm{diag}[0_{3\times 3}, -\hat\Omega],\quad
    \ad^*_{\xi_2} = \mathrm{diag}[-\hat\Omega_R, -\hat\Omega_L].
\end{gather*}
Finally, the matrix $C$ is given by
\begin{align*}
	C = \begin{bmatrix} 0 & 0 \\
		-Q_R & -Q_L \end{bmatrix}.
\end{align*}
The detailed developments of the above equations with explicit formulation of every term are presented in~\cite{csribb21, tejaswi2021geometric}. 
It is further shown that this model is consistent with the flight data of live Monarch butterflies detected by a motion capture system when its morphological parameters are chosen to mimic those of Monarch~\cite{sridhar2020geometric}.

\subsection{Wing and Abdomen Parameters}

As discussed above, the wing kinematics, or the attitude of wings relative to the body represented by $Q_R(t), Q_L(t)$ is formulated according to 1--3--2 Euler angles.
They are further parameterized by a specific model inspired by insect flapping as summarized below~\cite{berman2007energy}.
Let $f\in\Re$ be the flapping frequency in $\mathrm{Hz}$ and let $T=\frac{1}{f}$ be the period.
\begin{itemize}
	\item The flapping angle is given by
	\begin{align}
		\phi(t) & = \frac{\phi_m}{\sin^{-1} \phi_K}\sin^{-1}(\phi_K\cos(2\pi f t)) + \phi_0,\label{eqn:phi}
	\end{align}
	where $\phi_m\in\Re$ is the amplitude, $\phi_0\in\Re$ is the offset, and $0 < \phi_K \leq 1$ determines waveform shape.
	\item 
	The pitch angle is parameterized as
	\begin{align}
		\theta(t) = \frac{\theta_m}{\tanh \theta_C} \tanh( \theta_C \sin(2\pi f t + \theta_a)) +\theta_0,\label{eqn:theta}
	\end{align}
	where $\theta_m\in\Re$ is the amplitude of pitching, $\theta_0\in\Re$ is the offset, $\theta_C\in(0,\infty)$ determines the waveform, and $\theta_a\in(-\pi,\pi)$ describes phase offset. 
	\item 
	The deviation angle is given by
	\begin{align}
		\psi(t) = \psi_m \cos(2\pi \psi_N f t + \psi_a) + \psi_0,\label{eqn:psi}
	\end{align}
	where $\psi_m\in\Re$ is the amplitude, $\psi_0\in\Re$ is the offset, and the parameter $\psi_a\in(-\pi,\pi)$ is the phase offset. 
\end{itemize}
These are formulated for each of the left wing and the right wing. 
Thus the attitude, angular velocity, and acceleration of the wings can be reconstructed using \eqref{eqn:phi}--\eqref{eqn:psi}.

\subsection{Numerical Integrator}

The proposed imitation learning is model-free, i.e., instead of adopting the equations of motion \eqref{eqn:EL_xR} directly, optimal trajectories generated by \eqref{eqn:EL_xR} are utilized in construction of the control system. 
As such, it is critical to develop a faithful numerical integration scheme that reflects the dynamic characteristics of FWUAV properly. 
One particular challenge is that the rotation matrices representing the attitude of the body and wing kinematics evolve on a nonlinear Lie group.
It is well known that general purpose numerical integrators do not preserve the orthogonality of rotation matrices~\cite{haier2006geometric}.

More specifically, let the kinematics equations and the equations of motion be rearranged into
\begin{align}
    \dot g_1 &= g_1 \xi_1, \label{eqn:EL_num_0} \\
	\dot \xi_1 &= F(g_1, \xi_1, g_2, \xi_2), \label{eqn:EL_num}
\end{align}
where $ F $ can be constructed from \eqref{eqn:EL_xR}, and  $ g_2(t), \xi_2(t) $ are obtained from \eqref{eqn:phi}-\eqref{eqn:psi}.
As such, it corresponds to a time-varying ordinary differential equation on $\Re^3\times\SO3\times \Re^6$.

The common numerical integration techniques, such as the Runge Kutta method, rely on the addition operation. 
For instance, let $t$ be discretized into $\{t_0,t_1,\ldots\}$ with a fixed step size $h$. 
In the first-order Euler scheme, the attitude is updated by $ R(t_1) = R(t_0) + h R(t_0) \hat{\Omega}(t_0) $, which is not on $\SO3$ as addition is not closed in $\SO3$. 
Higher-order, adaptive integration schemes suffer from the same issue. 
Even though such standard RK methods have been utilized for optimal control implementation~\cite{tejaswi2021geometric}, it is observed that they are not robust.
When they are used to generate a significant amount of data for longer time durations, convergence is not guaranteed with the optimization setup described in the next subsection.

To address this, we utilize Lie group methods~\cite{iserles2000lie}, where the group elements are updated by the group operation to preserve group structures. 
In particular, a Crouch-Grossman method~\cite[Chapter IV.8]{haier2006geometric} is rearranged for \eqref{eqn:EL_num_0}--\eqref{eqn:EL_num} into
\begin{align}
    g_1(t_1) &= g_1(t_0) \prod_{i=1}^{s} \exp(h b_i \xi_1^i), \label{eqn:crg1} \\
	\xi_1(t_1) &= \xi_1(t_0) + h \sum_{i=1}^s b_i \zeta^i, \label{eqn:crgr}
\end{align}
where $ g_1^i\in\Re^3\times\SO3, \xi_1^i\in\Re^6, \zeta^i\in\Re^6 $ for $ i = 1, \dots, s $ are given by
\begin{align*}
	g_1^{i} &= g_1(t_0) \prod_{j=1}^{i-1} \exp(h a_{ij} \xi_1^j), \\
	\xi_1^{i} &= \xi_1(t_0) + h \sum_{j=1}^{i-1} a_{ij} \zeta^i, \\
	t^i &= t_0 + c_i h, \\
	\zeta^i &= f(g_1^i, \xi_1^i, g_2(t^i), \xi_2(t^i)).
\end{align*}
Here, $ a \in \Re^{s \times s}$ and $b, c \in \Re^s $ are real valued constants satisfying certain conditions for the method to guarantee an order of accuracy.
In particular, we use an explicit method of order $ 4 $ with $ s = 5 $ stages with the constant values in~\cite{jackiewicz2000construction}.
Moreover, the group element is updated by the group operation of $\Re^3\times\SO3$ in \eqref{eqn:crg1} to avoid the aforementioned issue of general purpose numerical integrators. 
As such, the numerical flow computed by \eqref{eqn:crg1}--\eqref{eqn:crgr} evolves on $\Re^3\times\SO3\times\Re^6$.
To speed up computation for simulation, \MATLAB coder is utilized to convert this numerical integrator to a `mex' file.
A sample result is shown here for a simulated trajectory where we observe that this method performs much better in comparison to a standard 4th order Runge Kutta method in terms of deviation from $ \SO3 $.
\begin{figure}[h!]
	\centering
	\includegraphics[width=0.8\linewidth]{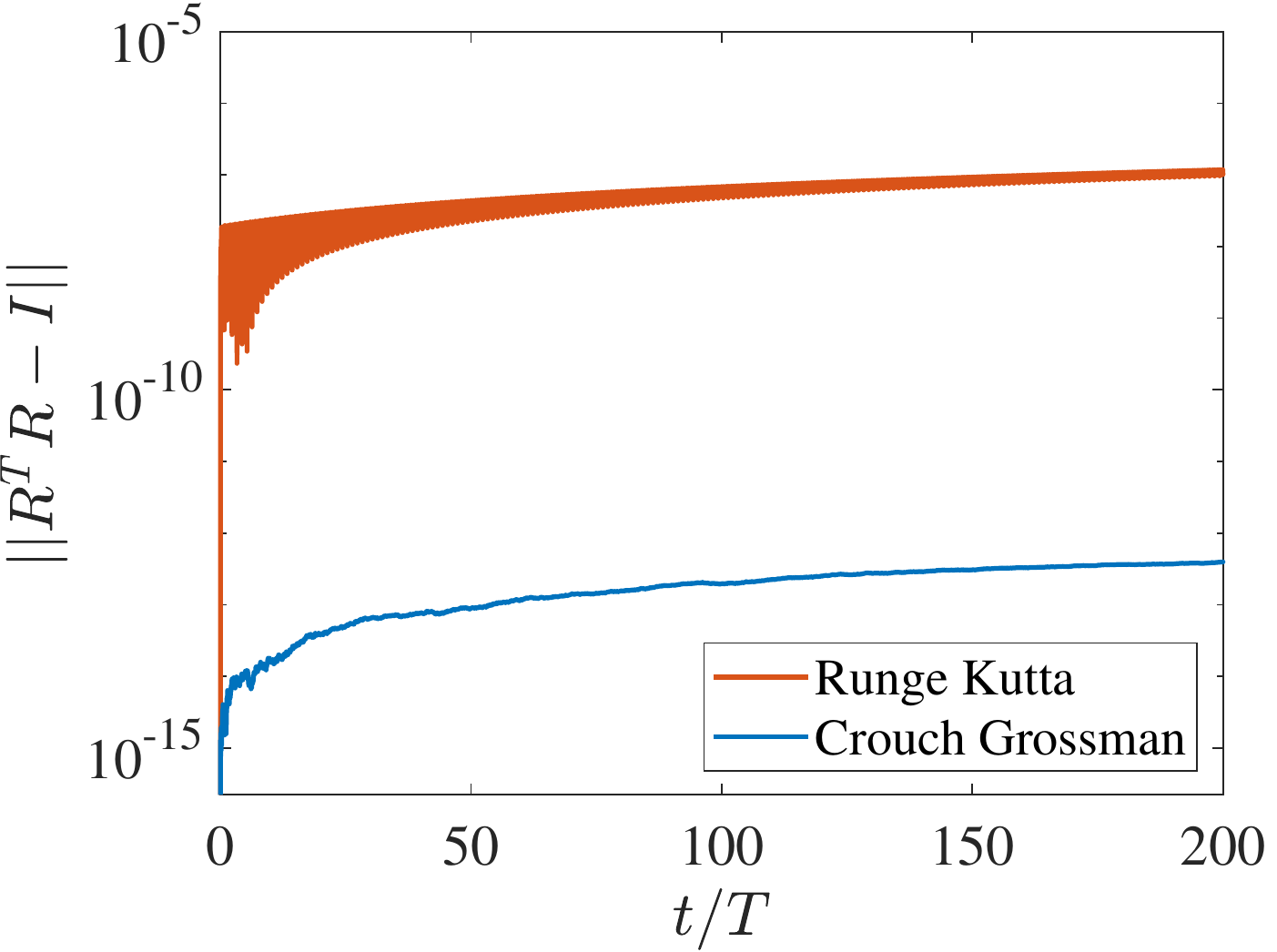}
	\caption{Comparison of orthogonality error}
	\label{fig:ortho_error}
\end{figure}

\subsection{Optimal Control of FWUAV}\label{sec:control}


The dynamic model presented here generates the trajectory of FWUAV for a given wing kinematics model, which is considered as a control input to FWUAV. 
The objective of the control system presented in this paper is to adjust the wing kinematics such that the motion of FWUAV asymptotically converges to a desired flight maneuver. 
This is addressed by generating a set of optimal trajectories for varying initial conditions, as presented in this section, which are utilized for imitation learning in the subsequent section. 

The first step is to construct a desired trajectory. 
However this is not trivial as the desired maneuver of FWUAV corresponds to a periodic orbit of the sophisticated dynamics. 
This has been formulated as a constrained optimization where the initial state, wing kinematics parameters, and flapping frequency are optimized to minimize the total energy while satisfying an equality constraint to enforce periodicity~\cite{tejaswi2021geometric}.
In this paper, we consider a hovering flight where the position and the attitude at the end of the flapping period match with the values at the beginning of the period as summarized in~\cite[Table 1]{tejaswi2021geometric}.

Next, controller parameters for optimization are defined as follows. 
If we directly utilize all of 24 wing kinematics parameters, the optimization problem will be ill-conditioned. 
To improve numerical properties and convergence of optimization, we select the following $N_\Delta = 6$ parameters that have dominant influence on the aerodynamic forces and torques:
\begin{equation}\label{eqn:params}
	\Delta = [\Delta \phi_{m_s}, \Delta\theta_{0_s}, \Delta \phi_{m_a}, \Delta\phi_{0_s}, \Delta\theta_{0_a}, \Delta\psi_{0_a}].
\end{equation}
The above controller parameters are defined as follows. 
Let $\Delta$ denote the deviation from the wing kinematics of the reference trajectory.
For example, $ \Delta \phi_{m,R} = \phi_{m,R} - \phi_{m,R,d} $ represents the change of the flapping amplitude of the right wing from its desired value. 
The second-level subscript $s$ corresponds to the symmetric or the average change between the left and the right wing, i.e., $\Delta\phi_{m_s} = \frac{1}{2}(\Delta\phi_{m,R}+\Delta\phi_{m,L})$, and the subscript $a$ is for the asymmetric change, i.e., $\Delta\phi_{m_a} = \frac{1}{2}(\Delta\phi_{m,R}-\Delta\phi_{m,L})$.
All other variables are defined in a similar manner.
The effects of these parameters on each component of the aerodynamic force and moment are summarized in~\cite[Table 2]{tejaswi2021geometric}.
Instead of changing these parameters at each time step, we divide the flapping period $[0, T] $ into $N_s = 10 $ steps at which the values of $\Delta$ are specified.
In between these points, $\Delta(t)$ values are interpolated in a piecewise linear manner.
An additional constraint $ \Delta(0) = \Delta(T) = 0 $ is imposed to ensure that the desired trajectory is periodic.

Let $\mathbf{x}(t) = (g_1(t),\xi_1(t)) = (x(t), R(t), \dot x(t),\Omega(t))$ be the state of FWUAV representing the translational and rotational dynamics.
The desired reference trajectory for the hovering flight is denoted by $ \mathbf{x}_d(t) = (x_d(t), R_d(t), \dot x_d(t), \Omega_d(t)) $.
The above parameters are optimized to ensure $ \mathbf{x}(t) \to \mathbf{x}_d(t)$, i.e., the trajectory asymptotically converges to the periodic orbit when $\mathbf{x}(0)\neq \mathbf{x}_d(t)$.
The objective function for optimization is the discrepancy between the desired trajectory and the controlled trajectory, measured by
\begin{align}
	J = \sum_{i=1}^{N_p} W_i \sqrt{\sum_j (W_{\mathbf{x}_j} (\mathbf{x}_j(t_i) - \mathbf{x}_{d_j} (t_i)))^2},
\end{align} 
where the time steps are $ t_i = i \times T/N_s $ over a horizon $N_p$, and the subscript $j$ iterates over each component of each element of $\mathbf{x}$.
Thus state errors at time $ t_i $ are combined in the inner sum after weighing by a factor $ W_{\mathbf{x}_j} $ across each component.
This ensures that all elements are scaled by their own physical characteristics leading to a single measure of state error.
Next, these errors are added over time in the outer sum across a horizon of $ N_p $ steps, after being weighed by another factor $ W_i $.
These weights are designed to increase over time so that the final state errors have more importance.

While solving this problem we adopt the procedure of model predictive control (MPC) using the `fmincon' solver from \MATLAB.
That is, we obtain control parameters over a prediction horizon of two time periods ($ N_p = 20 $) while the actual controller is implemented only for the first period ($ N_s = 10 $).
At the end of this period, the process of optimization is repeated.
The detailed procedure for optimization is presented in~\cite{tejaswi2021geometric}.


\section{Constrained Imitation Learning}\label{sec:IL}

The optimization procedure illustrated in the previous section is not suitable for real-time implementation,
as the time required for optimization is greater than the flapping period by several orders:
a few minutes are required for optimization while a single flapping period is on the scale of $ 0.1 $ seconds.
In this section, we adopt the framework of imitation learning, where an online feedback control scheme is constructed from a set of optimal trajectories obtained offline. 
Popular imitation learning schemes, namely behavior cloning, DAgger and DART are introduced, and a new constrained imitation learning is proposed. 

\subsection{Behavior Cloning}

One of the most straightforward approaches of imitation learning is utilizing supervised learning to construct a controller that is trained to reproduce the above optimal control trajectories for a given state.
For instance, a multilayered neural network can be trained over a set of optimal control inputs generated in the previous section.
Here we exploit the fact that the desired maneuver is a periodic orbit.
More specifically, the input of the neural network is the state of FWUAV at the beginning of a flapping period and the output is wing kinematics parameters over the entire flapping period. 
This implies that the feedback mechanism of comparing the actual state to the desired state is activated periodically at the beginning of each period.
This is sensible as the same state errors at two distinct flapping phases imply different dynamic characteristics, and it reduces the dimension of the input for the neural network substantially. 
Numerical examples considered in this paper indicate that checking the state error once per a period yields satisfactory results for stabilization.
However, this approach is readily extended to incorporate state errors in other phases of the flapping period. 

To generate the training data set, the above optimization problem is solved repeatedly for varying initial conditions. 
We select random initial errors in the form of $ \Delta \mathbf{x}(0) = [\Delta x(0), \Delta R(0), \Delta \dot x(0), \Delta \Omega(0)] \in \Re^{12} $, defined by
\begin{gather*}
	\Delta x = x - x_d,\quad \Delta R = \frac{1}{2} (R_d^T R - R^T R_d)^\vee, \\
	\Delta \dot x = \dot x - \dot x_d,\quad \Delta \Omega = \Omega - R^T R_d \Omega_d.
\end{gather*}
Each part of this error vector is sampled from the uniform sphere in $ \Re^3 $,
and it is scaled by a factor to ensure that all states are varied in a similar physical level.
For each initial error, the preceding optimization problem is solved to construct the optimal control parameters $u\in\Re^{60}$, which is the values of $\Delta\in\Re^6$ defined in \eqref{eqn:params} over $N_s=10$ points in a flapping period. 
The training data are composed of $N$ pairs of the corresponding $(\Delta \mathbf{x}(0), u)$, which also include $N_0$ pairs of $(\Delta \mathbf{x}(0) = 0, u=0)$ to promote that zero state error results in zero control.

\begin{figure}[b]
	\centering
	\includegraphics[trim={1.5cm 0cm 1.5cm 0cm},clip,width=\linewidth]{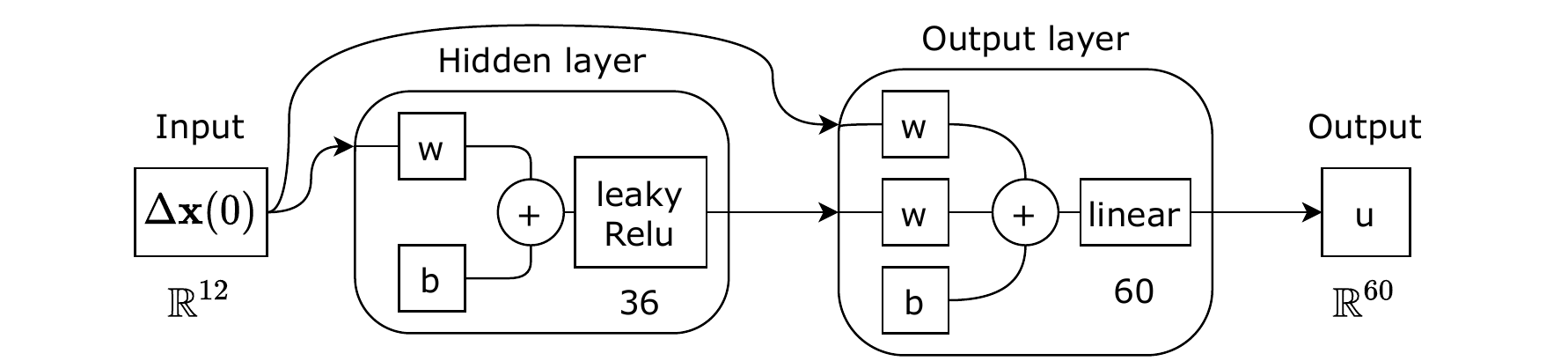}
	\caption{Neural network architecture}
	\label{fig:control_net}
\end{figure}

Next a neural network is formulated so that control can be implemented without actually solving the computationally expensive optimization problem for arbitrary initial errors.
We use a simple cascade-forward architecture with a single hidden layer composed of $ 36 $ leakyRelu neurons (see \Cref{fig:control_net}).

Let the input matrix be $ X \in \Re^{12 \times N} $, whose columns $ \braces{X_k, k = 1\dots N} $ correspond to the values of $ \Delta \mathbf{x}(0) $.
Similarly the output matrix $Y \in\Re^{60\times N}$ is constructed by the optimal control parameters. 
The input-output relation of the neural network is represented by
$
    \hat{Y} = f(X, \theta),
$
where $ \theta \in \Re^{N_\theta} $ are the parameters of the neural network composed of weights and biases, and $ \hat{Y} $ is the predicted output.
The parameters $\theta$ are trained to minimize a loss function, $ L(\hat{Y}, Y) $ which is often chosen as the mean squared error in regression.
The training algorithm utilized in this paper is Bayesian regularization backpropagation~\cite{foresee1997gauss} which exhibits highly desirable generalization to untrained inputs.

The detailed implementation and the corresponding controlled trajectories are presented in \Cref{sec:results}.
It turns out that the resulting performance is not satisfactory.
The control input constructed by this network leads to instabilities after transient responses.
Following an initial convergence of the state errors with respect to the reference trajectory, the controlled trajectories start to diverge as the number of flapping increases.
This is not surprising as the presented approach based on supervised learning, referred to as behavior cloning, does not perform well in practice.
It is because the particular states encountered in the controlled trajectory may not be well represented by any of the optimal trajectories in the training data. 
The mismatch of the distributions between the training data set and the actual controlled trajectory accumulates over time inevitably.

\subsection{DAgger}

The issues of behavior cloning can be mitigated by increasing the size of the training data 
so that any state in the controlled trajectory is sufficiently close to an optimal trajectory in the data set.
However this will increase the computational cost substantially in terms of both trajectory optimization and neural network training.
Alternatively, a popular benchmark algorithm, referred to as DAgger addresses the problems of behavior cloning by integrating online learning and iterative training~\cite{ross2011reduction}.
It involves interaction with the expert during training as summarized below.

\begin{algorithm}
	\begin{algorithmic}[1]
		\REQUIRE $ (X, Y), N_i \in \mathbb{Z} $
		\FOR {$ i = 1$ \TO $N_{i}$}
		\STATE $ \theta^i = \argmin_\theta \{L(\hat{Y}, Y) \mid \hat{Y} = f(X, \theta)\} $
		\STATE Execute the trained policy to construct on-policy trajectories 
		\STATE Add these new states, $ X_k $ into the input set $ X $
		\STATE Obtain optimal control for new states through optimization, $ Y_k $, and add them to the output set $ Y $
		\ENDFOR
		\ENSURE $\theta^{N_i}$
	\end{algorithmic}
	\caption{DAgger}
	\label{alg:DAgger}
\end{algorithm}

\Cref{alg:DAgger} describes the DAgger approach as applied to our control problem.
With the given data set, the neural network model is trained at Step 2 as in behavior cloning.
Then in Step 3, the trained neural network is executed over several periods to get new controlled trajectories.
The offline optimal control presented in \Cref{sec:control} is utilized for these new states to obtain the corresponding optimal control inputs,
which are added to the training dataset.
This procedure is repeated for $ N_i $ iterations to improve the performance.

In short, training is repeated with the new data set obtained from on-policy, online trajectories. 
As the particular states encountered in the controlled trajectories are constantly added to training, the aforementioned issue of the distribution mismatch is directly addressed.
However, this approach is computationally expensive as the size of the training data increases and the time-consuming trajectory optimization should be solved repeatedly. 
In the perspective of imitation learning, the expert (represented by trajectory optimization in this paper) should be available for interaction during the learning process.

\subsection{DART}

Another interesting approach known as DART~\cite{laskey2017dart} proposes noise injection into the expert's policy while collecting data.
This is done off-policy unlike DAgger thus avoiding visits to potentially unsafe states during training, leading to a decrease in computation.
However it can provide corrective guidance to the controller since it enlarges the boundary of expert's distribution.
The actual value of noise to be utilized is determined by minimizing the difference between distributions of the optimal control expert and the trained policy.

The application to our regression problem with a Gaussian noise injected supervisor is mentioned in Algorithm \ref{alg:DART}.
Firstly, Step 1 corresponds to an initial training with data obtained without noise.
This is a slight modification to the original method since we empirically observe an increase in performance when we do so.
Then in Step 3, we estimate the current noise parameter which is the covariance matrix, $ \hat{\mathbf{\Sigma}} $, based on available samples.
Here, we represent the current neural network policy as $ \pi_{\theta^{i}} $ and the expert policy as $ \pi_{\theta^*} $.
Later this value is scaled in Step 4 to include the anticipated final policy error through a parameter, $ \alpha_d $.
Now we can implement the noise-injected expert which is just the optimal controller from \Cref{sec:control} with the additional randomness.
Finally, the neural network model is trained at Step 7 on the aggregated dataset.
These iterations are repeated for $ N_i $ number of times.

\algsetup{
	indent=2em,
	linenodelimiter=.,
}
\begin{algorithm}
	\linespread{1.}\selectfont
	\begin{algorithmic}[1]
		\REQUIRE $ (X, Y), N_i \in \mathbb{Z}, \alpha_d \in \Re $
		\STATE $ \theta^0 = \argmin_\theta \{L(\hat{Y}, Y) \mid \hat{Y} = f(X, \theta)\} $
		\FOR {$ i = 1$ \TO $N_{i}$}
		\STATE $ \hat{\mathbf{\Sigma}}^i \approx \frac{1}{N} \sum_k (\pi_{\theta^{i-1}}(X_k) - \pi_{\theta^*}(X_k)) (\pi_{\theta^{i-1}}(X_k) - \pi_{\theta^*}(X_k))^T $
		\STATE $ \displaystyle \mathbf{\Sigma}^i = \frac{\alpha_d}{N \trs{\hat{\mathbf{\Sigma}}^i}} \hat{\mathbf{\Sigma}}^i $
		\STATE Collect data using the optimal control expert with injected noise : $ \mathcal{N}(\pi_{\theta^*}, \mathbf{\Sigma}^i) $
		\STATE Add these input output pairs to the dataset $ (X, Y) $
		\STATE $ \theta^i = \argmin_\theta \{L(\hat{Y}, Y) \mid \hat{Y} = f(X, \theta)\} $
		\ENDFOR
		\ENSURE $\theta^{N_i}$
	\end{algorithmic}
	\caption{DART}
	\label{alg:DART}
\end{algorithm}

Thus DART addresses some of the drawbacks of DAgger through an off-policy procedure of adding noise to the controller.
It is still able to minimize the mismatch between distributions of the expert and the trained policy.
However, it needs to interact with the optimal control framework which is computationally expensive.

\subsection{Constrained Imitation Learning}~\label{sec:COIL}

Now we propose constrained imitation learning to address these issues. 
It turns out that the main challenge for behavior cloning is its inability to keep state error near zero after an initial convergence.
Due to the limited number of optimal trajectories in the training data, the controller is not accurate specifically near the desired trajectory with zero state error.
Even though the controller performs reasonably well during the initial transient period with large errors, it is unable to maintain the boundedness of the small error over a longer period of time. 

This has been mitigated by augmenting the training data set with the additional pairs of $(\Delta \mathbf{x}(0)=0,u=0)$. 
But, it does not resolve the issue completely.
Another alternative workaround would be enforcing the constraint of zero input to zero output directly on the neural network structure, by eliminating all of the bias terms in \Cref{fig:control_net}.
However, this method drastically reduces the learning capabilities of the neural network and it results in poor generalization.

In this paper, we propose a new imitation learning scheme inspired by the recent studies on supervised learning with constraints~\cite{detassis2020teaching}.
To improve long-term convergence, the zero input, zero output constraint is incorporated in the training process.
Instead of varying the neural network parameters to learn the target and to satisfy the constraint simultaneously,
the learning target is adjusted to mitigate the computational challenges in constrained supervised learning. 
The foremost advantage is that it does not require the solution of trajectory optimization for new states online, and therefore, the size of the training data is fixed.

More specifically, the neural network training for the proposed constrained imitation learning is formulated into the following optimization problem:
\[
    \theta^* = \argmin_\theta \braces{L(\hat{Y}, Y) \mid \hat{Y} = f(X, \theta), \text{ and } f(0, \theta) = 0}.
\]
The proposed procedure to address this optimization is summarized at \Cref{alg:iter}.
\algsetup{
	indent=2em,
	linenodelimiter=.,
}
\begin{algorithm}[h!]
	\linespread{1.}\selectfont
	\begin{algorithmic}[1]
		\REQUIRE $ (X, Y), \alpha \in \Re, N_i \in \mathbb{Z} $
		\STATE $ \theta^0 = \argmin_\theta \{L(\hat{Y}, Y) \mid \hat{Y} = f(X, \theta)\} $ \\
		\hfill \COMMENT {initial training}
		\STATE $ \hat{Y}^0 = f(X, \theta^0) $
		\FOR {$ i = 1$ \TO $N_{i}$}
		\STATE $ \theta_{z_0} = \argmin_\theta \{L(Z_0, (1-\alpha)Y + \alpha \hat{Y}^{i-1}) \mid Z_0 = f(X, \theta)\} $
    \STATE $ \mathbf{F} \approx \frac{1}{N} \sum_{k} \nabla_\theta \log p_{\theta_{z_0}}(Y_k | X_k) \nabla_\theta \log p_{\theta_{z_0}}(Y_k | X_k)^T $
		\STATE $ \theta_z = \argmin_\theta \frac{1}{2} \norm{\theta - \theta_{z_0}}_\mathbf{F}^2, \quad f(0, \theta) = 0 $ \\
		\hfill \COMMENT {constrained optimization}
		\STATE $ Z^i = f(X, \theta_z) $ \hfill \COMMENT {target adjustment}
		\STATE $ \theta^{i} = \argmin_{\theta} \{L(\hat{Y}, Z^i) \mid \hat{Y} = f(X, \theta)\} $ \\
		\hfill \COMMENT {unconstrained training}
		\STATE $ \hat{Y}^i = f(X, \theta^i) $
		\ENDFOR
		\ENSURE $\theta^{N_i}$
	\end{algorithmic}
	\caption{Constrained Imitation Learning (COIL)}
	\label{alg:iter}
\end{algorithm}

First, there is the initial training step which is implemented without considering any constraints.
This part is composed of Steps 1 and 2, and it is equivalent to behavior cloning.
Next, the iterative parts of Steps 3 through 10, are motivated by the main algorithm in \cite{tejaswi2022iterative} and consist of the combination of target adjustment and unconstrained training.
Essentially the target $Y$ is adjusted into a new target $Z$ such that unconstrained training with respect to $Z$ automatically satisfies the given constraint. 

Steps 4-7 correspond to  the process of finding the alternative target $ Z $ that is more feasible for the constraint. 
Initially in Step 4, the neural network is trained such that its output becomes closest to $ (1-\alpha)Y + \alpha \hat{Y} $ for $ \alpha \in [0, 1] $.
This is on the line connecting the ideal output $Y$ and the output $\hat Y^{i-1}$ adjusted for the constraint in the prior iteration.
As such, this has the net effect of adjusting the neural network parameters to find the compromise between emulating the ideal output (smaller $\alpha$) and satisfying the constraint (larger $\alpha$).

The next part of Steps 5 and 6 directly addresses the constraint via nonlinear constrained optimization. 
In Step 6, the neural network parameters $\theta_z$ are optimized to enforce the zero input, zero output constraint while minimizing the parameter deviation from the value $\theta_{z_0}$ obtained in Step 4.
The motivation for the objective function $\|\theta-\theta_{z_0}\|^2_{\mathbf{F}}$ is to ensure consistency of the resulting control policy. 
In other words, we wish that the optimal control system represented by the neural network is not drastically altered while enforcing the constraint of $f(0,\theta)=0$.

More specifically, let $ \pi_\theta $ represent the Gaussian policy obtained by the current value of neural network parameters $\theta$, i.e., $\pi_\theta(Y|X)\sim\mathcal{N}(f(X,\theta), \mathbf{\Sigma})$ for a covariance $\mathbf{\Sigma}\in\Re^{60\times 60}$.
Similarly, $\pi_{\theta_z}$ is constructed from $\theta_{z_0}$.
The difference between those two control policies can be measured by the KL-divergence approximated by
\begin{equation}
    D_{KL} (\pi_\theta || \pi_{\theta_{z_0}}) \approx \frac{1}{2}(\theta-\theta_{z_0})^T \mathbf{F} (\theta-\theta_{z_0}),
\end{equation}
where $ \mathbf{F} \in \Re^{N_\theta \times N_\theta} $ is the Fisher information matrix (FIM)~\cite{pascanu2013revisiting,schulman2015trust} defined by
\begin{align}
    \mathbf{F} = \mathrm{E}_{\pi_{\theta}} \bracket{\nabla_\theta \log \pi_\theta(Y | X) \nabla_\theta \log \pi_\theta(Y | X)^T}.\label{eqn:F}
\end{align}
For the Gaussian distribution, the gradient term in \eqref{eqn:F} can be evaluated by
\begin{align}
	\nabla_\theta \log \pi_\theta(Y | X) &= \nabla_\theta \parenth{-\frac{1}{2} \norm{Y - f(X, \theta)}_{\mathbf{\Sigma}^{-1}}^2} \nonumber \\
                                       &= (\nabla_\theta f(X, \theta)) \mathbf{\Sigma}^{-1} (Y - f(X, \theta)). \label{eqn:grad_logp}
\end{align}
The gradient $\nabla_\theta f$ in \eqref{eqn:grad_logp} is computed through backpropagation, and it is readily available for any deep learning software library. 
During numerical implementation, the arithmetic mean in \eqref{eqn:F} is replaced by the sample mean with each element of the training data, as shown at Step 5,
and the covariance matrix can be chosen as $ \mathbf{\Sigma} = I $ for our purposes~\cite{pascanu2013revisiting}.

The role of the Fisher information matrix is critical in Step 6.
If we directly optimize a simple square error term, $ \|\theta - \theta_{z_0}\|^2 $, then it takes a lot of iterations to converge.
This is because the sensitivity of $f(X,\theta)$ with respect to $\theta$ varies greatly. 
In other words, some parameters change the output of the neural network drastically while others are not strongly correlated to the output.
The matrix $\mathbf{F}$ properly scales the neural network parameters to have more uniform effects on the output, thereby improving computational properties of optimization.

Once we obtain the adjusted parameter $\theta_z$, the corresponding target adjusted for the constraint is computed in Step 7. 
In the remaining steps, the neural network is retrained for the adjusted target, and these iterations are repeated $N_i$ times. 
The desirable feature is that the neural network training in Step 4 and Step 8 do not consider the constraint explicitly. 
As such, these steps can be performed with any supervised learning technique. 

The key difference in this procedure from \cite[Algorithm 1]{tejaswi2022iterative} (which we will refer to as $ RC $) is the explicit form of the constraint.
More specifically,
\begin{itemize}
	\item The zero input, zero output constraint that is enforced here is a function of the parameters of the neural network, $ f(0, \theta) = 0 $.
	But, the type of constraint considered in $ RC $ is on the predicted output matrix, $ \hat{Y} $, so that it belongs to a feasible set
	.
	\item Essentially, the constraint here simplifies to a relation in the network parameters, $ \theta $, whereas in $ RC $ it becomes a relation among the elements of the predicted output, $ \hat{Y} $.
	\item So the infeasible adjustment step from $ RC $ has to be modified into multiple steps leading to the constrained optimization in Algorithm \ref{alg:iter}.
	That is, firstly in Step 4 the network parameters are estimated after training on $ (1-\alpha)Y + \alpha \hat{Y} $.
	Next, the constraint is enforced in Step 6 after the Fisher information matrix $ \mathbf{F} $ is estimated using these obtained parameters.
	\item Moreover, there is no feasible adjustment step of $ RC $ since the constraint here will not be exactly satisfied.
\end{itemize}

This Algorithm \ref{alg:iter} is referred to as constrained imitation learning (COIL).
Compared with DAgger and DART, the most distinctive benefit is that we do not have to solve the cumbersome trajectory optimization during the iteration. 
In other words, there is no need to communicate with the expert to acquire an additional optimal trajectory. 
This also does not increase the size of the training data constantly. 
While there are two neural network training steps per an iteration, compared with a single training in the other two algorithms, each training corresponds to minor adjustments for the adjusted targets within the existing data.
Thus, it can be solved more efficiently compared with the training in DAgger and DART involving new input and output pairs. 
In short, the zero input, zero output constraint is carefully incorporated into the neural network training without increasing the computational load drastically. 
This is illustrated by numerical examples in the next section.

\section{Numerical Simulation}\label{sec:results}

In this section, we present numerical properties of all of behavior cloning, DAgger, DART and COIL applied to a FWUAV model presented in~\cite{csribb21, tejaswi2021geometric}.
We focus on evaluating the performance through the initial convergence rate and the ultimate boundedness, compared against the required computation time.
As it is not feasible to investigate stability properties with Lyapunov analysis, we perform an exhaustive numerical study where the controlled trajectories are computed for numerous initial conditions within a prescribed domain.
This is to ensure reasonable performance for any new initial condition that is not close to training data.

More specifically, motivated by the definition of ultimate uniform boundedness in~\cite{kha02}, we identify positive constants $\gamma, t_T, b$ satisfying
\begin{gather}
	\norm{\mathbf{x}(t) - \mathbf{x}_d(t)}_{W_\mathbf{x}} \le \norm{\mathbf{x}(0) - \mathbf{x}_d(0)}_{W_\mathbf{x}} e^{-\gamma t}, \forall\ 0 \le t \le t_T, \nonumber\\
	\norm{\mathbf{x}(t) - \mathbf{x}_d(t)}_{W_\mathbf{x}} \le b,\quad \forall\ t \ge t_T
\end{gather}
where $ \norm{\mathbf{x}}_{W_\mathbf{x}} = \sqrt{\sum_i (W_{\mathbf{x}_i} \mathbf{x}_i)^2}$ is a weighted two norm.
Here $\gamma$ corresponds to the initial exponential convergence rate, and $b$ is the ultimate bound.
Next, $ t_T $ is the first time to reach the ultimate bound.

\subsection{Controller Performance}
The above values are determined by simulating 12291 trajectories initiated in the set of $ \norm{\mathbf{x}(0) - \mathbf{x}_d(0)}_{W_\mathbf{x}} \le 1 $ for each method.

\begin{figure}[h!]
	\centering
	\includegraphics[width=\linewidth]{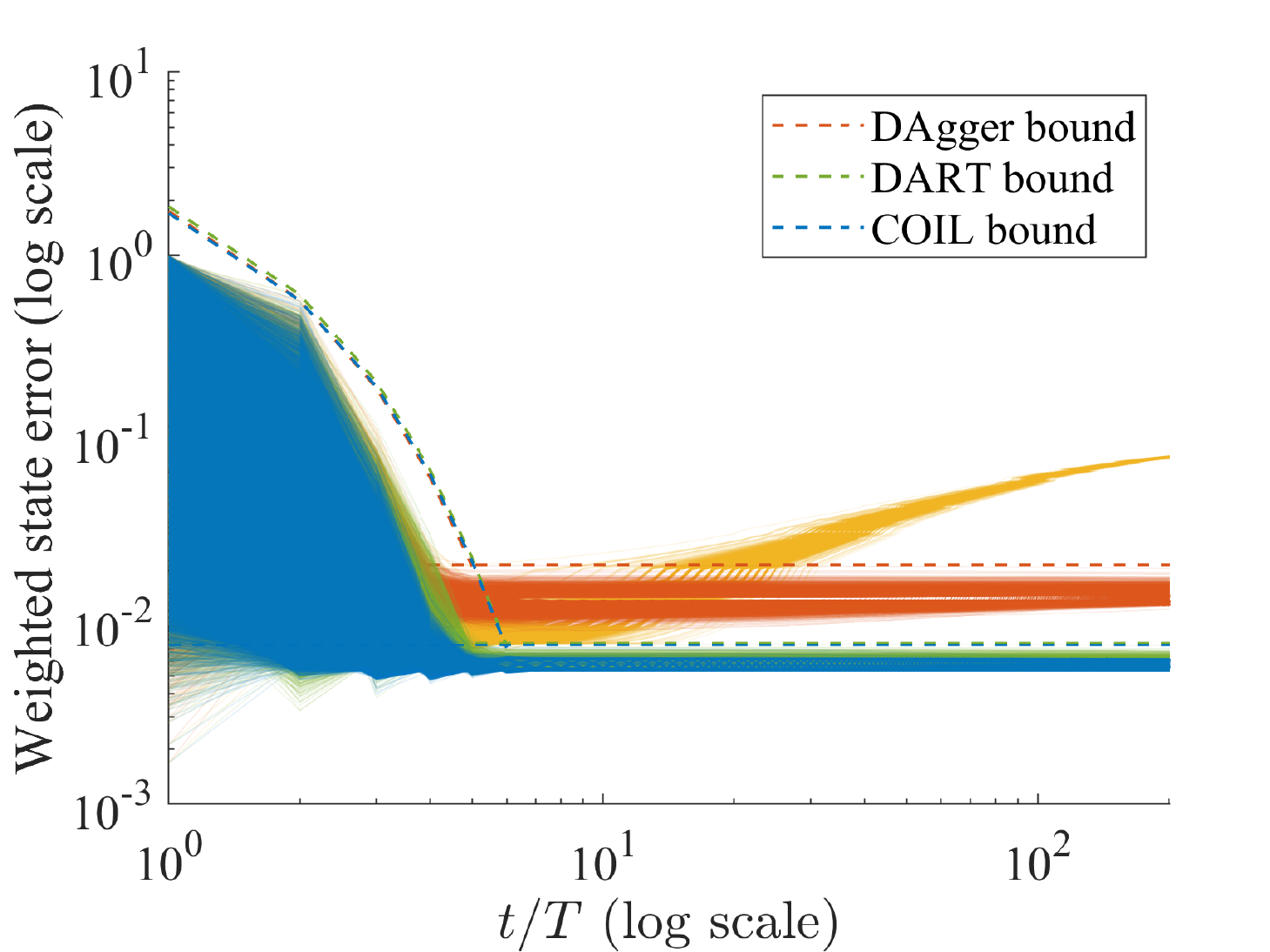}	
	\caption{Comparison of controller performance : Behavior Cloning (yellow), DAgger (orange), DART (green), COIL (blue); dashed lines represent the ultimate bounds while the initial decay is characterized by a bounding exponential function}
	\label{fig:perf_comp}
\end{figure}
\paragraph{Behavior Cloning}
Behavior cloning is implemented with a neural network that has $ N_\theta = 3408 $ parameters (Figure \ref{fig:control_net}), and the size of the training dataset is $ N = 1587 $ including $ N_0 = 300 $ pairs of zero points.
The corresponding convergence of the tracking error is illustrated in \Cref{fig:perf_comp} for all initial conditions.
It is observed that there is a non-negligible increase of the error after about 50 flapping periods, implying an ultimate bound can not be characterized for the given simulation period.

\paragraph{DAgger}
This is implemented according to Algorithm \ref{alg:DAgger} with $ N_i = 5 $ iterations.
At Step 3, the trained neural network is executed over $ 5 $ time periods to obtain around $ 150 $ new controlled trajectories.
However the controller failed to make the error converge.
Therefore, we had to increase the number of neurons in the hidden layer from $36$  to $ 60 $ (resulting $ N_\theta = 5160 $) and also  the number of initial training data to $ N = 3049 $.
The resulting tracking error is presented in \Cref{fig:perf_comp}.
The issue of behavior cloning is addressed to attain the ultimate bound of $ b = 0.0204 $. 
However, this is achieved with more training data and a larger neural network.

\paragraph{DART}
Next, Algorithm \ref{alg:DART} corresponding to Disturbances for Augmenting Robot Trajectories (DART) is implemented with $ N_i = 5 $ iterations again.
The scaling factor in Step 4 is taken to be $ \frac{\alpha_d}{N \trs{\hat{\mathbf{\Sigma}}^i}} = 10^{-4} $.
This relatively small number is picked since in our case, even a minor change in the control value obtained from expert led to a large perturbation in the states.
Then in Step 5, data is collected using the noisy optimal controller over $ 5 $ time periods resulting in $ 150 $ new values.
Unlike DAgger the smaller network structure with $ N_\theta = 3408 $ is adequate here, but the size of initial training data in Step 1 still had to be slightly increased to $ N = 2012 $.
It can be observed from \Cref{fig:perf_comp} that DART realized a better ultimate bound of $ b = 0.0076 $.

\paragraph{Constrained Imitation Learning}
Finally, the proposed COIL is implemented by Algorithm \ref{alg:iter} with the same number of iterations $ N_i = 5 $ as DAgger.
The trade-off parameter between the ideal and current labels is taken to be $ \alpha = 0.75 $ which gives a little more weight to constraint adjustment.
The network structure and the training data are identical to those of behavior cloning, i.e., $N_\theta = 3408$ and $N=1587$.
As presented in \Cref{fig:perf_comp}, COIL successfully achieved ultimate boundedness with a lower value of $ b = 0.0074 $ as compared to DAgger, and similar to that of DART, both of which rely on more training data. 

\subsection{Results}

\begin{table}
	\caption{Numerical comparison}
	\label{tab:comp}
	\begin{center}
		{\begin{tabular}{c|cccc}
				\toprule
				\multirow{2}{*}{Algorithm} & 
				\multirow{2}{*}{\shortstack[c]{BC}}&
				\multirow{2}{*}{\shortstack[c]{DAgger}} &
				\multirow{2}{*}{\shortstack[c]{DART}} &
				\multirow{2}{*}{\shortstack[c]{COIL}}
				\\ \\
				\midrule
				Computation time (min) & 3.8 & 132.22 & 131.08 & 44.53  \vspace*{1mm}\\
				Ultimate bound, $b$ & N/A & 0.0204 & 0.0076 & 0.0074 \vspace*{1mm} \\
				Initial decay rate & 1.1296 & 1.1168 & 1.1046 & 1.0944 \\
				\bottomrule
		\end{tabular}}
	\end{center}
\end{table}

These results are summarized in Table \ref{tab:comp}, 
with computation time when implemented in \MATLAB with Intel(R) Xeon(R) Silver 4216 CPU.
While behavior cloning requires the shortest computation time, its performance is not desirable as presented above.
DAgger achieves ultimate boundedness, but it requires the most computation time,
which is caused by time-consuming trajectory optimization solved in every iteration (Step 5 in \Cref{alg:DAgger}).
Also, the more complex network architecture leads to a drastic increase in the computation time. 
Meanwhile, DART avoids visiting unsafe states in training and performs better compared to DAgger in terms of the final error.
On the other hand, the proposed COIL is able to obtain the best ultimate bound and computation time while maintaining a similar initial bounding decay rate. 
This directly illustrates the desirable features of COIL addressing the issue of distribution mismatch of behavior cloning without excessive additional computation required in DAgger and DART.

On the other hand, a neural network can theoretically approximate any function given enough amount of data and model complexity.
For instance, if the initial dataset dimension is increased to more than $ 10000 $ (an order higher), even Behavior Cloning will perform decently.
However, this will be very expensive since the MPC expert needs to generate this amount of data.
COIL just forces the neural network model to include a zero input to zero output constraint and improve convergence by reshaping the policy in a stabilizing manner.
The exact value of this constraint measured as $ \norm{f(0, \theta)}_2 $ for each of DAgger, DART and COIL are $ 0.1849, 0.0618, 0.0007 $ respectively.
On the other hand, the mean squared error in training the network are $ 0.0062, 0.0063, 0.0083 $ respectively.
Of course, COIL has the best value in terms of constraint satisfaction since we explicitly include it in the iterative algorithm, leading to a trade off to a higher value of MSE.
Ultimately the controller's impact in testing is important, where it is competitive with a standard benchmark like DART with the advantage of lower computational time.

The idea of using neural network policies is to be able to implement in real-time as compared to optimal control.
Now the average time taken by the MPC expert is over $ 100 $ seconds if run on a serial process (single worker/ CPU).
On the other hand, the inference time of the NN policy for a given state error is less than $ 10^{-4} $ seconds when run without parallelization.
In the real setup, it would be necessary to get new control values at the start of each trajectory time period which is around $ 0.085 $ seconds (flapping frequency of $ 11.75 $ Hz).
Hence it can be reasoned that using a fast controller represented by the neural network is preferable.

\begin{figure}
	\centerline{
		\subfigure[Linear velocity error]{
			\includegraphics[width=0.5\linewidth]{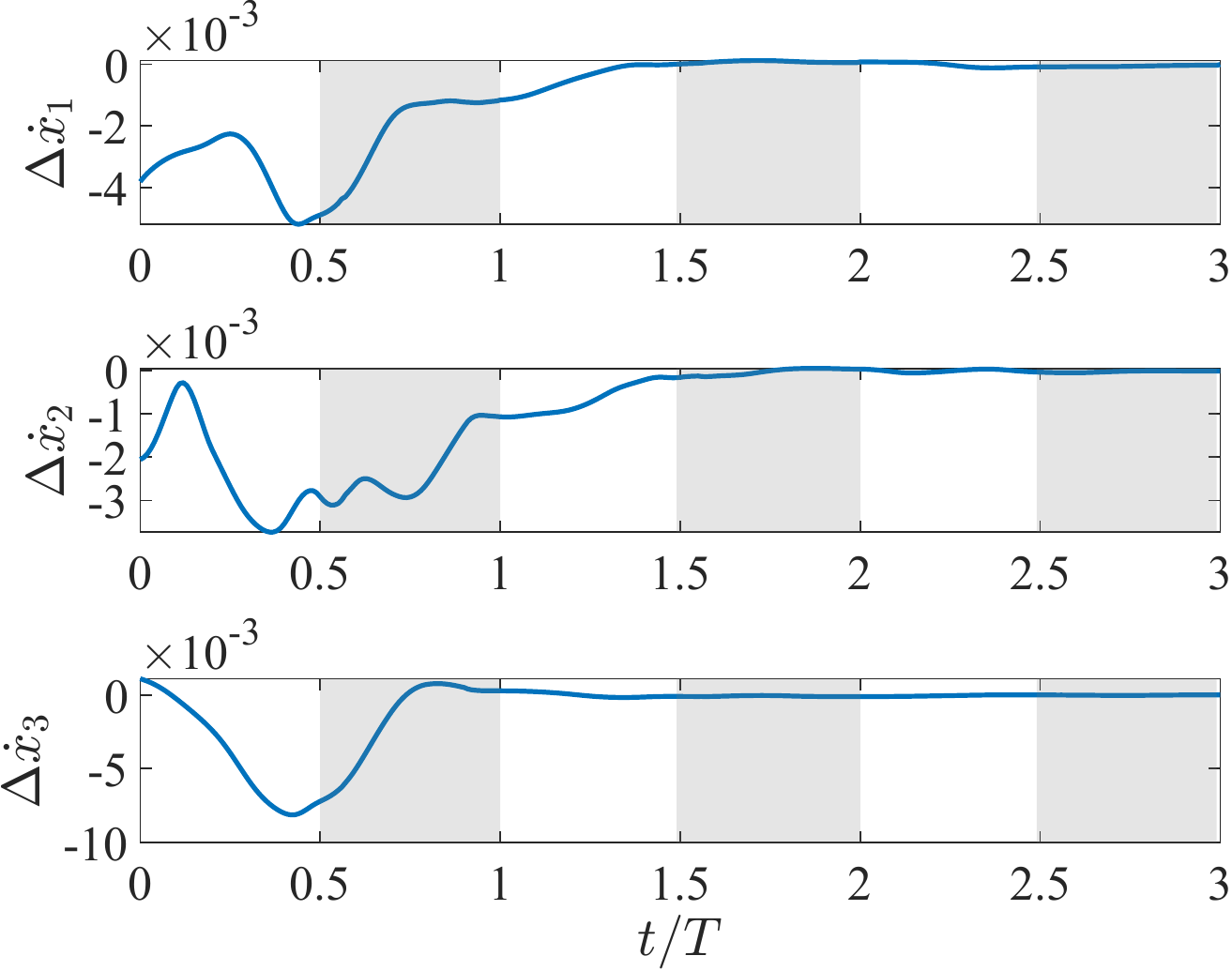}
		}
		\hfill
		\subfigure[Angular velocity error]{
			\includegraphics[width=0.5\linewidth]{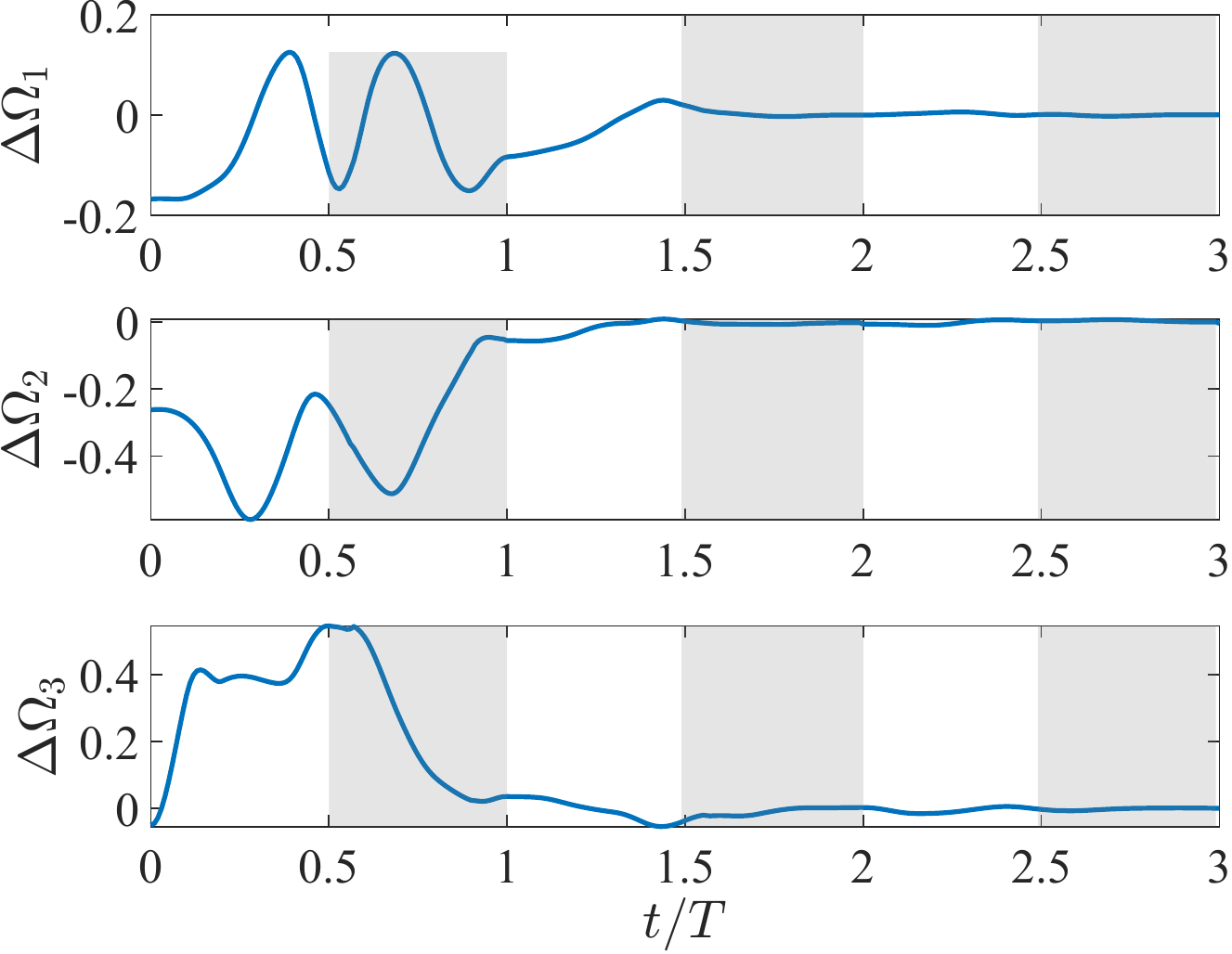}
		}
	}
	\centerline{
		\subfigure[Position and attitude error]{
			\includegraphics[width=0.5\linewidth]{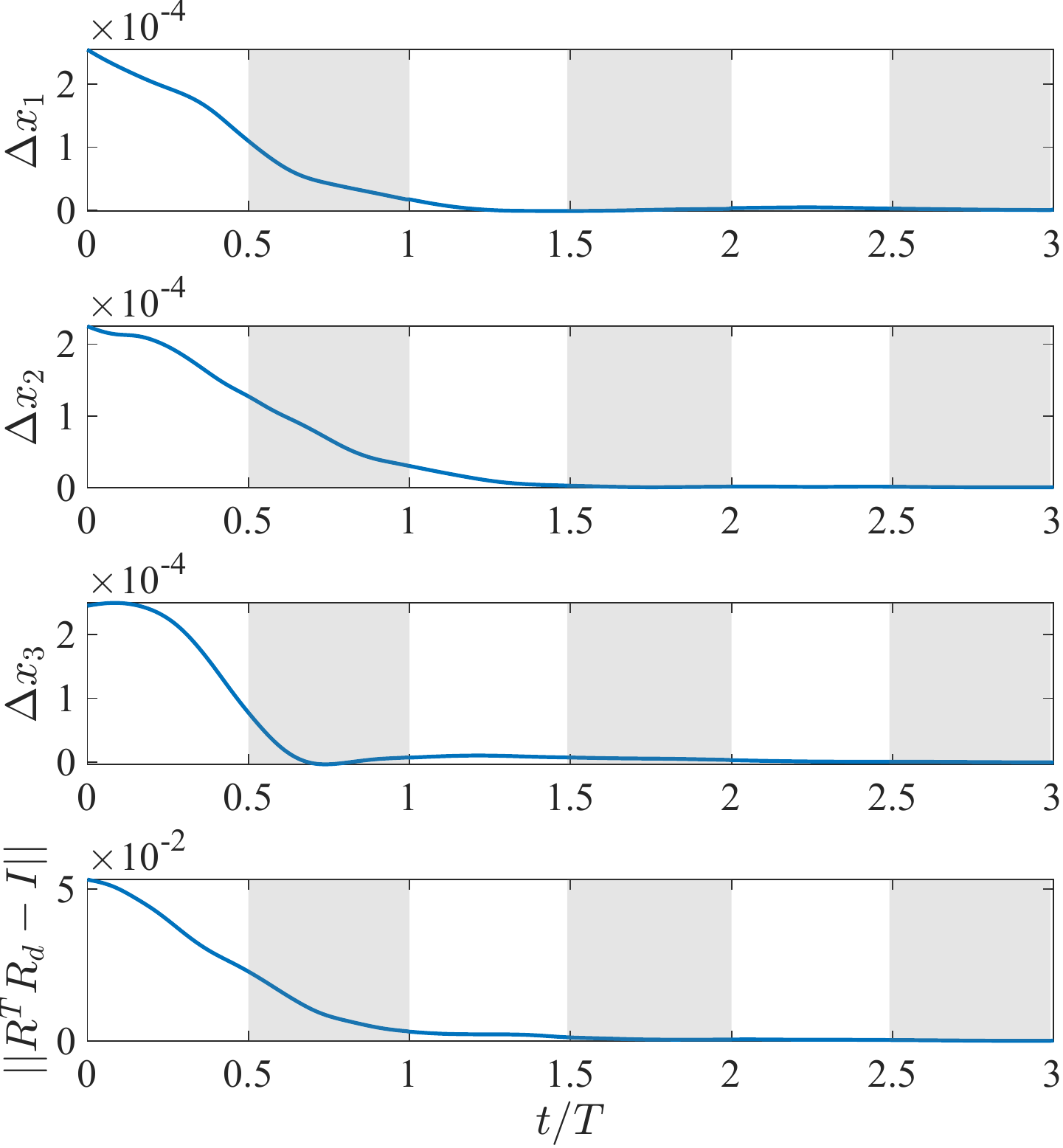}
		}
		\subfigure[Control inputs]{
			\includegraphics[width=0.475\linewidth]{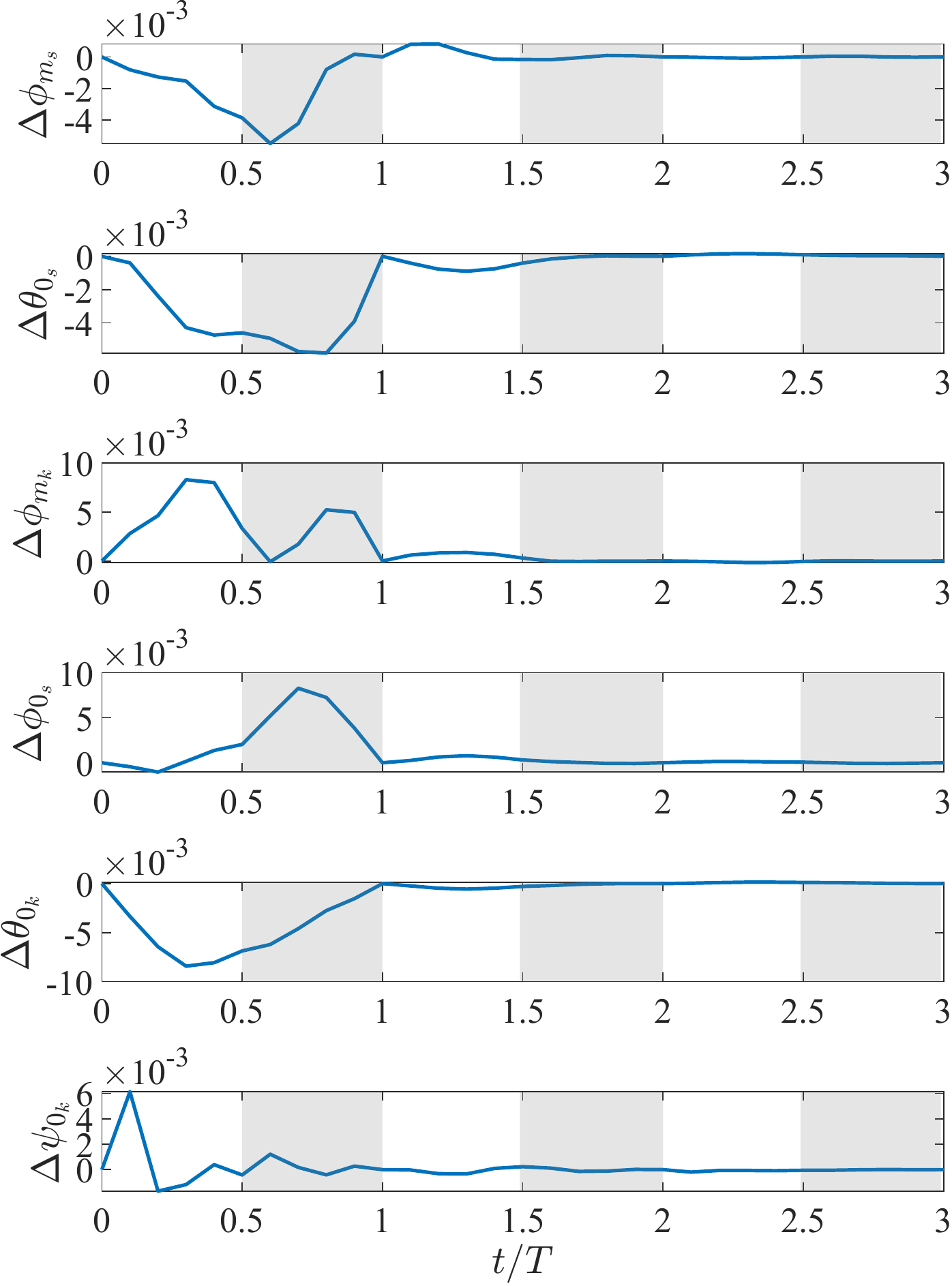}
		}
	}
	\caption{
		Optimal trajectory errors and control parameters}
	\label{fig:hover_control}
\end{figure}

As an illustration, the tracking error in each of position, velocity, and attitude for a single particular trajectory is presented in \Cref{fig:hover_control}, along with COIL control inputs.
This illustrates that the hovering flight of FWUAV is successfully stabilized without the common averaging or linearization for the coupled longitudinal and lateral dynamics.

\subsection{Robustness to Noise}

\begin{figure}[h!]
	\centering
	\includegraphics[width=\linewidth]{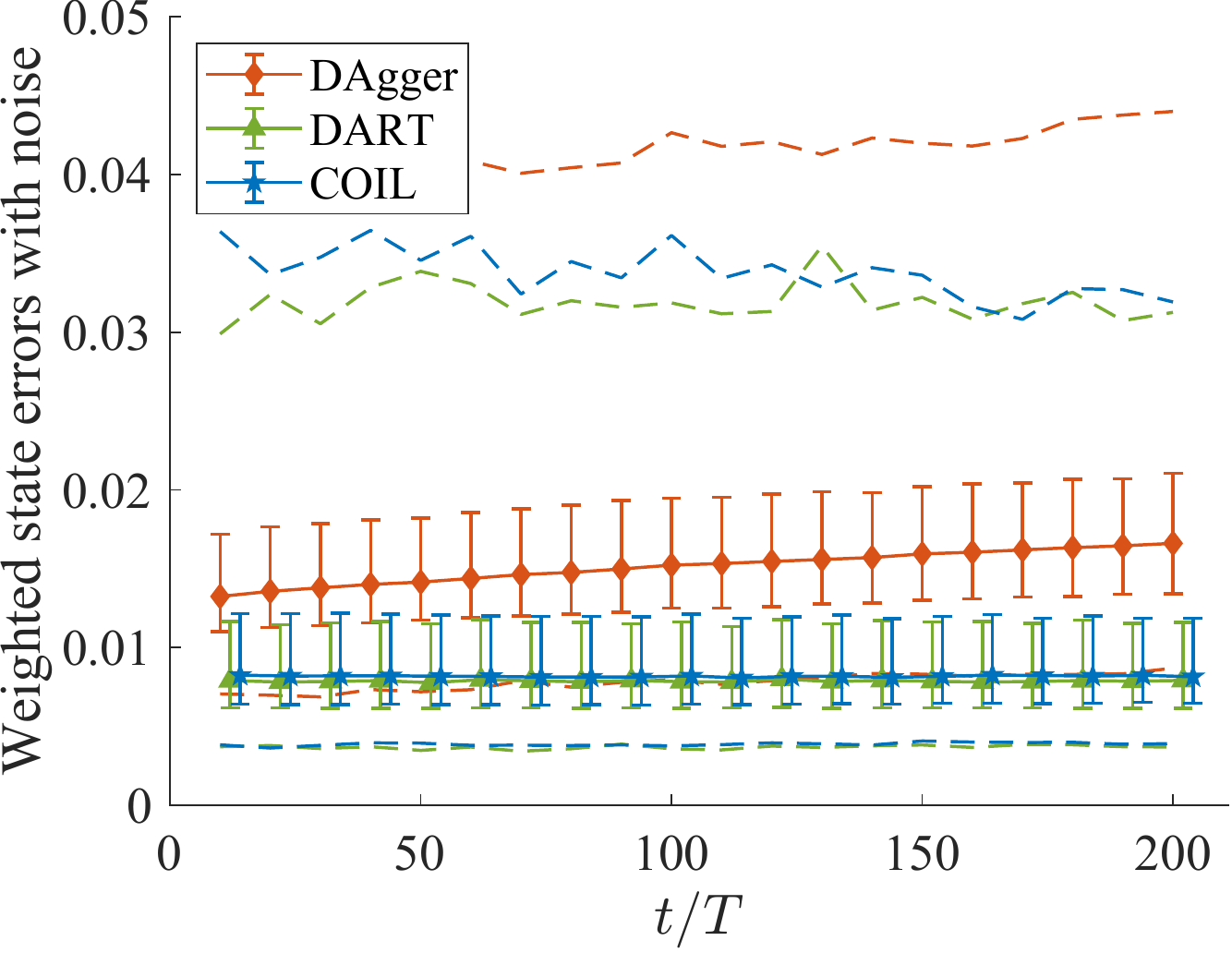}	
	\caption{Comparison of controller performance with measurement noise : box plots represent $ 0.25 $, $ 0.5 $ (median), $ 0.75 $ quartiles of errors; dashed lines correspond to maximum and minimum values}
	\label{fig:noise_comp}
\end{figure}

In real world systems it is essential for the policy to be robust to noise in the state estimation or modeling.
Here, we consider the case when there is measurement noise in the controller setup.
Random Gaussian noise corresponding to a weighted scale of $ 10^{-3} $ is added to the actual state value before being passed to the trained neural network.
Multiple trajectories with similar initial conditions as in \Cref{fig:perf_comp} are obtained in simulation using the controllers from DAgger, DART and COIL.

The corresponding tracking errors for $ \frac{t}{T} > 10 $ are presented in the form of summary statistics in \Cref{fig:noise_comp}.
Values for the first $ 10 $ periods are skipped since the trend remains very close to \Cref{fig:perf_comp} when the magnitude of state errors is large.
It can be observed that DAgger implemented as before does not work ideally since the trend of errors is increasing.
It will require more iterations in Algorithm \ref{alg:DAgger} to perform well enough.
Meanwhile, both DART and COIL seem to be equally robust to noise in that state errors slightly decline over time.
Thus the proposed COIL algorithm is promising in its application to handle noise in real scenarios.


\section{Conclusions}
This paper presents imitation learning for a flapping wing aerial vehicle inspired by Monarch butterflies,
where a set of optimal trajectories computed offline is considered as an expert demonstration to be emulated. 
A new constrained imitation learning is proposed to achieve ultimate boundedness followed by exponential convergence without relying on additional trajectory optimization, thereby improving computational efficiency substantially. 
This is successfully applied to the nonlinear dynamics of a flapping wing aerial vehicle. 
A potential future direction is utilizing visual input directly in sensorimotor control since it is challenging to estimate the complete state of flapping wing aerial vehicle in practice.

\bibliographystyle{IEEEtran}
\bibliography{root}

\end{document}